\documentclass[twoside,leqno,twocolumn]{article}

\usepackage[letterpaper]{geometry}
\usepackage{ltexpprt}
\usepackage{amsmath,amssymb,amsfonts}

\usepackage[bold,full]{complexity}

\usepackage{algorithm}   
\usepackage{algorithmic}  

\usepackage{graphicx}
\usepackage{booktabs}
\usepackage{subfig}
\usepackage{adjustbox}
\usepackage{float}
\newcommand{\cnp}{\textbf{NP}}

\DeclareMathOperator*{\argmax}{arg\,max}

\usepackage{hyperref}
\usepackage{amsmath}
\usepackage{subfig}
\usepackage{esvect}
\usepackage{url}

\usepackage{color}
\usepackage{amssymb}
\usepackage{empheq} 
\usepackage{amsmath}
\usepackage{commath}
\usepackage{indentfirst}
\usepackage{amsfonts}
\usepackage{xcolor}

\begin{document}

\title{\Large A Graph-Based Approach for Active Learning in Regression}

\author{Hongjing Zhang\thanks{Department of Computer Science,
University of California, Davis, hjzzhang@ucdavis.edu,
davidson@cs.ucdavis.edu} \and S.~S. Ravi\thanks{Biocomplexity Institute
and Initiative, University of Virginia and University at Albany -- SUNY,
ssr6nh@virginia.edu} \and
Ian Davidson\footnotemark[1]}

\date{}

\maketitle


\fancyfoot[R]{\scriptsize{Copyright \textcopyright\ 2020 by SIAM\\
Unauthorized reproduction of this article is prohibited}}





\begin{abstract} \small\baselineskip=9pt 
\noindent Active learning aims to reduce labeling efforts by
selectively asking humans to annotate the most important data points
from an unlabeled pool and is an example of human-machine interaction.
Though active learning has been extensively researched for
classification and ranking problems, it is relatively understudied
for regression problems. Most existing active learning for regression
methods use the regression function learned at each active learning
iteration to select the next informative point to query. This
introduces several challenges such as handling noisy labels, parameter
uncertainty and overcoming initially biased training data. Instead,
we propose a feature-focused approach that formulates both sequential
and batch-mode active regression as a novel bipartite graph
optimization problem. We conduct experiments on both noise-free and
noisy settings. Our experimental results on benchmark data sets demonstrate
the effectiveness of our proposed approach.\newline
\end{abstract}


\section{Introduction}
\label{introduction}
Supervised learning methods assume a large well-annotated training
set. However, in many real-world applications, labeled data is often
difficult and expensive to obtain, but we may have a large pool of
unlabeled data. In such settings, active learning can be used where
the machine learns an initial model from the labeled data, and then
repetitively asks a human to annotate instances from the unlabeled
pool.

Active learning has been extensively studied for classification and
is most useful in settings where there are limited training annotations
\cite{cohn1996active,tong2001support,nguyen2004active,dasgupta2008hierarchical,qian2013fast,huang2014active}.
However, there are several limitations with existing active learning:
1) Most works focus on classifications with relatively few works
for regression. 2) Most works in active regression are model-focused;
they use the learned regression function to select the next query
point. Regression is an under-studied problem by the research
community but extensively used in practice due to its ease of use
and interpretability.


The limitations of being model focused are subtle but important.
For example, as the query depends on the model if the previous point
queried is incorrectly labeled by the domain expert, then it can
produce undesirable results including even increasing model error.
Most active learning for regression studies
\cite{burbidge2007active,sugiyama2009pool,cai2013maximizing,sabato2014active,riquelme2017online}
assume that the annotations are accurate and focus on selecting the
most informative point to query based on the newly learned function
at each iteration. However, the labelers in the real world usually
make noisy annotations, especially for regression tasks
\cite{malago2014online, zhang2015active}.
Furthermore, the model-based methods could be highly biased when
initial labeled data is limited as we discuss in Section
\ref{sec:experimental}. We use the term \emph{challenging regression
settings} to denote the regression problems where initial labeled
data is limited and noisy due to human annotations.

In this paper, we propose a feature-focused active sampling strategy
that tackles active learning for regression \emph{without} using
the regression function. Instead, we formulate the active learning
problem as a \emph{bipartite graph optimization problem} to reduce
uncertainty in the unlabeled instances. One set of nodes corresponds to labeled
points while the other to unlabeled points. The choice of the best points
to move from the unlabeled set to the labeled set is analogous 
to the classic $k$-Median
and $k$-Center problems. Even though these problems are known to be computationally intractable in general, we
can adopt a known efficient approximation algorithm for the $k$-Median problem to
solve the batch mode query active learning problem. To demonstrate
the versatility of our approach, we explore classic regression
settings, including classic linear regression and polynomial
regression. A core challenge is how to estimate label uncertainty,
and we propose a $L_1$ based measure and show that optimizing this measure
is equivalent to optimizing the model's uncertainty upper bound. Our
new approach has been evaluated on several datasets, and it is
shown to outperform many state-of-the-art methods (Section
\ref{sec:experimental}).
The main contributions and novelty of this paper are summarized as follows.
\begin{itemize}
\item We formulate a feature (not model) focused active learning
algorithm as a bipartite graph optimization problem. We show that this
problem is, in general, computationally intractable but observe that
good approximation algorithms exist
(Section \ref{sec:complexity} and Theorem \ref{thm:mmmd_complexity}).
\item We develop a $L_1$-based measure that upper bounds uncertainty
(Theorem \ref{thm:bound}) and use it to create edge weights in our
optimization problem (Section \ref{sec:algorithms}).
\item We create both sequential query and batch mode query algorithms
for our formulation (Algorithms \ref{alg:1}, \ref{alg:2}). Although
we derive our uncertainty measure from the linear regression, we
show its versatility by using it with other popular regression formulations.
\item We experimentally demonstrate the effectiveness of our proposed
algorithm and the tightness of our bound in both \emph{normal} and
\emph{challenging} regression settings (Section \ref{sec:experimental}).
\end{itemize}

The rest of the paper is organized as follows. In Section
\ref{sec:related_work} we discuss related work. In Section
\ref{sec:prob_def} we formulate our active learning algorithm in a
general bipartite graph optimizing problem. In Section \ref{sec:complexity}
we connect out formulation with well-known graph problems and provide
complexity results for the general version of our formulation. In
Section \ref{sec:algorithms} we introduce both sequential and batch
mode query algorithms which minimize the total uncertainty upper
bound for linear regression and then extend them to polynomial
regression for complex data. In Section \ref{sec:experimental} we
demonstrate our algorithm's performance on various domains with
noise-free and noisy annotations. Section \ref{sec:conclusion}
provides concluding remarks. 

\section{Related Work}
\label{sec:related_work}
There are three main categories of methods
for querying unlabeled points in active learning. 
The first category is finding the most \emph{informative} or
discriminative example for the current model. Cohn et al. proposed
an algorithm \cite{cohn1996active} that minimizes the learner's
error by minimizing its variance to reduce generalization error,
assuming a well-specified model and an unbiased learning function
and data. Burbidge et al. propose an adaptation of Query-By-Committee
\cite{burbidge2007active} in active learning for regression. Sugiyama
introduced a theoretically optimal active learning algorithm
\cite{sugiyama2009pool} that attempts to minimize the generalization
error in the pool based setting. Yu provided passive sampling
heuristics \cite{yu2010passive} to shrink the space of candidate
models based on the samples' geometric characteristics in the feature
space. Cai et al. presented a sampling method in regression, which
queries the point leading to the largest model change
\cite{cai2013maximizing}. Although our approach is based on uncertainty
sampling which looks for informative points, we are looking not for
a single point with the most uncertainty, 
but for one that can bring overall uncertainty reduction.

The second category is to find the most \emph{representative} points
for the overall patterns of the unlabeled data while preserving the
data distribution \cite{yu2006active,chattopadhyay2013batch}. In
particular, clustering for better sampling representative points has
been explored \cite{nguyen2004active,dasgupta2008hierarchical}.
Considering representative points gives better performance when
there are no or very few initially labeled data. 
However, such approaches have two
major weaknesses: their performance heavily depends on the quality
of clustering results and their efficiency (in comparison with methods
that use informative points) will degrade as the number
of labeled data points increases. 
The idea of a representative point is considered in our approach as we
look for overall uncertainty reduction, but we also consider finding an
informative point based on our uncertainty measure.

The third category of methods considers \emph{informativeness
and representativeness} simultaneously. Existing work which was
motivated by this goal achieved excellent performance
\cite{huang2014active,wang2015querying}. Our work belongs in this
category but is different from previous work as we select valuable
points based on the instances and not on the learned regression function.

Standard active learning methods usually assume that there is 
an oracle that can provide accurate annotations for each query.
In the real world, the annotations could be noisy. There has been
much work on active learning for classification under different
noise models and with diverse labelers
\cite{yan2011active,zhang2015active,huang2017cost}. Instead of
making assumptions on the specific noise type and diverse labelers,
our method seeks a model-free sampling strategy to make our
active learning model less vulnerable to the potential noise
in annotated data.

\section{Our Idea and Problem Definition}
\label{sec:prob_def}
In this section, we study how to generate active queries for
regression methods only using \emph{feature space} properties of
the points. We calculate the uncertainty of each unlabeled point
and then ask the oracle to label a subset of points 
that provides the maximum reduction in the \emph{overall} uncertainty.

Assume that we have a small labeled set ${L} = {{\{ (x_i, y_i) \}}^{l}_{i=1}}$
with $l$ instances, where $y_i$ is the label of the $i$th instance
$x_i$, and a larger pool of unlabeled data ${U} = {{\{ x_j \}}^{n}_{j
= l + 1}}$ with $n - l$ instances. At each iteration of active
learning, the algorithm selects a subset of unlabeled instances and
queries the oracle (i.e., a domain expert) to obtain their labels. 
We aim to choose a subset
of points that will reduce the uncertainty by the maximum amount.
Uncertainty is reduced
both \emph{directly} (the selected unlabeled points are given labels)
and \emph{indirectly} (the uncertainty of the remaining unlabeled
points can be further reduced based on newly labeled points).

This query decision is made based on an evaluation function $Q$,
which measures the total reduction of uncertainty after querying
the new unlabeled points. We first formulate our proposed algorithmic
framework as a bipartite graph optimization problem. We then propose
a simple yet effective way to calculate the function $Q$ and finally
present our active learning algorithm.


Consider a graph where each node represents a data point. In pool-based
active learning, the nodes are partitioned into two groups $L$ and $U$
corresponding to labeled and unlabeled points respectively.
For each node $i \in U$, we add an edge to a node $j \in L$
such that the point corresponding to $j$ 
\emph{most} reduces the label uncertainty of the point corresponding to $i$.
The weight of this edge is the uncertainty of $i$'s labeling.
This creates a weighted bipartite graph $G([U,L],W)$ between $U$
and $L$ with edge set $W$.


Given $G([U,L],W)$, our goal is to reduce the total estimated
uncertainty $H$ by choosing $S_u \subset U$ to add to $L$, where $H$
is defined as follows:
\label{prob:1}
\begin{align*}
H = \sum_{i \in U, j \in L} W_{i,j}
\end{align*}
However, our aim is not merely to choosing the most uncertain points. 
We also want the chosen subset to reduce the uncertainty of others points
in $U$ (i.e., those that were not chosen). We then have a new graph
$G'([U-S_u,L+S_u],W')$ and the total uncertainty that associated
with all unlabeled points is now changed to:
\begin{align*}
H' = \sum_{i \in U-S_u, j \in L+S_u} W'_{i,j}
\end{align*}
Letting
$Q(S_u)$ denote the total uncertainty reduction from $H$ to $H'$, 
our goal is to choose $S_u$ with size $k$ that maximizes $Q(S_u)$:
\begin{equation*}
\argmax_{S_u} Q(S_u) =\sum_{i \in U, j \in L} W_{i,j} - \sum_{i \in U-S_u, j \in L+S_u} W'_{i,j}
\end{equation*}

\begin{figure}[t]
 \vskip -0.1in
\small
 \centering
 \includegraphics[scale=0.38]{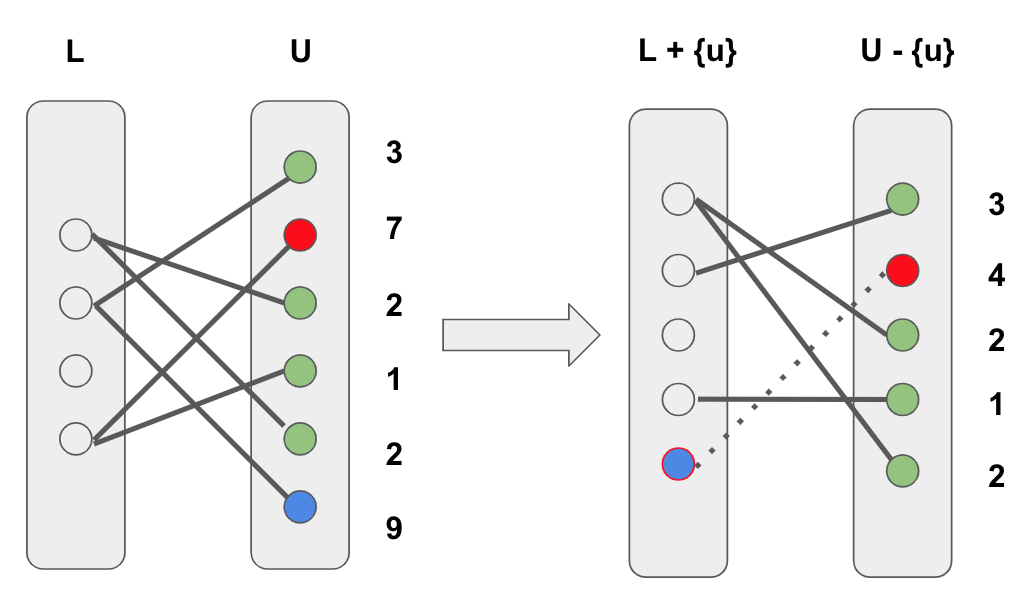}
 \caption{\small{A toy example for our bipartite graph view of active
 learning. Left: Each unlabeled point has an edge connected to its
 closest labeled point whose uncertainty is given on the right
 side. Right: By labeling/moving the bottom point in $U$ to
 $L$, we directly reduce its uncertainty (from $9$ to $0$) and
 indirectly reduce the uncertainty of another point (2nd from the
 top) from $7$ to $4$.}}
 \vskip -0.2in
 \label{fig:bipartite}
\end{figure}

We illustrate our basic intuition with a toy example in Figure
\ref{fig:bipartite} where $S_u$ has only one unlabeled point
$u$. Querying unlabeled point $u$ will move it from $U$ to $L$ and
change the edges in the graph. After the querying, 
the uncertainty of $u$ has been removed since $u$ has been
labeled and its corresponding uncertainty changes from $9$ to $0$.
This is an example of directly reducing uncertainty. However, $u$
is the nearest neighbor of an unlabeled point and this unlabeled
point's uncertainty is reduced from $7$ to $4$. This is an example
indirectly reducing uncertainty. The estimated total reduction in uncertainty
is $Q(u) = 9 + (7-4) = 12$. Our approach aims to find a set $S_u$ that maximizes
the estimated total reduction in uncertainty. 

\section{Complexity Results}
\label{sec:complexity}

In this section, we first formulate a general version of our bipartite
graph optimization problem and establish its complexity.  
This theoretical analysis guides us in developing approximation algorithms for active learning which will be introduced in Section \ref{sec:algorithms}. This section can be skipped on the first reading of the paper.

A bipartite graph $G([U, L], W)$ is a \textbf{nearest neighbor
bipartite graph} (or NN-BG) if it satisfies the following two
conditions. 
\begin{itemize} 
\item A distance
value $d(x,y) \geq 0$ is given for each pair of nodes $x, y$ 
where $x \in L$ and $y \in U$.
\item For each node $y \in U$, $W$ contains \emph{exactly}
one edge $\{x,y\}$, where $x \in L$ is a \emph{nearest neighbor}
of $y$ among all the nodes in $L$ (i.e., for each node $z \in L$,
$d(x,y) \leq d(z,y)$).  
\end{itemize}
Thus, in any NN-BG $G([U, L], W)$, each node in $U$ has exactly 
one edge incident on it; therefore, $|W| = |U|$. 


Given an NN-BG $G([U, L], W)$, suppose we move a non-empty subset of nodes $S \subseteq U$ from $U$ to $L$. After this modification, we can find the nearest neighbor for each node $y \in U-S$ and obtain another NN-BG denoted by $G'([U-S, L \cup S], W')$. We consider the following two problems involving such modifications of NN-BGs.


\noindent
\textbf{I. Modification to Minimize the Maximum\newline Distance} (MMMD)

\noindent
\underline{\textsf{Instance:}} Given an NN-BG $G([U, L], W)$, an integer $k \leq |U|$ and a non-negative number $\beta$.

\noindent
\underline{\textsf{Question:}} Is there a subset $S \subseteq U$
such that (i)~ $|S| \leq k$ and (ii)~ in the NN-BG $G'([U-S, L \cup
S], W')$, the distance value on each edge $\{x,y\} \in W'$ is $\leq$
$\beta$?


\noindent
\textbf{II. Modification to Minimize the Total\newline Distance} (MMTD)

\noindent
\underline{\textsf{Instance:}} Given an NN-BG $G([U, L], W)$, an
integer $k \leq |U|$ and a non-negative number $\sigma$.

\noindent
\underline{\textsf{Question:}} Is there a subset $S \subseteq U$
such that (i)~ $|S| \leq k$ and (ii)~ in the NN-BG $G'([U-S, L \cup
S], W')$, the sum of the distances over all the edges in $W'$ is $\leq$
$\sigma$?


We can show that both MMMD and MMTD are \cnp-complete even when the
distance function is a metric. The detailed proofs are shown in the
supplementary material.

\begin{theorem}\label{thm:mmmd_complexity}
Problems MMMD and MMTD are \cnp-complete even when the distance function is a metric.
\end{theorem}

\section{Putting it All Together - Our Algorithms}
\label{sec:algorithms}
Here, we first show how to calculate an uncertainty upper bound
for each unlabeled point and then translate it as the weight of an edge 
in the bipartite graph $G([U, L], W)$. Further, we propose two
algorithms for sequential and batch-mode active learning, which are
different approximation algorithms for the bipartite graph
optimization problem. Finally, we extend our active learning strategy
for linear regression to polynomial regression, which suits more
complex data sets.

\subsection{Motivating the Use of $L_1$ Distance as an Uncertainty Measure}
We now show that the optimum way to calculate the edge weights ($W$)
is equivalent to constructing a $L_1$ distance measure. Uncertainty
sampling is one of the most popular algorithms for active learning
in classification \cite{settles2012active}. For example, in
margin-based methods like SVMs \cite{tong2001support} we can calculate
the distance between the unlabeled point and the decision boundary
as an uncertainty measure. Similarly, in probabilistic models
\cite{settles2008multiple}, we can calculate the entropy of unlabeled
points for uncertainty. Generally speaking, this method computes a
measure of the classifier's uncertainty for each example in the
unlabeled set and then returns the most uncertain one. However, the
measurement of uncertainty in regression is not as straightforward
as for classification problems \cite{yu2010passive}.

We define an intuitive uncertainty measure as follows. Let the data
be $d$-dimensional. Suppose the current regression model (learned with
${L}$) predicts for $x$: 
$$f(x) = \sum_{i=1}^{d} w_i x_i + b$$ 
and the ideal model (learned with $L \cup U$) predicts: 
$${f^*}(x) = \sum_{i=1}^{d} w_i^* x_i + b^*$$ 
We can rewrite the
target value of an arbitrary unlabeled point $x_u$ by adding and
subtracting the $f$ and $f^*$ value for a labeled point $x_l$. Note
that the positive/negative bias terms ($b$) cancel each other out.

\begin{equation} \label{eq1}
 \resizebox{.9\hsize}{!}{$f(x_u) = f(x_l)+f(x_u)-f(x_l) = f(x_l) + \sum_{i=1}^{d} w_i (x_{u_i} - x_{l_i})$}
\end{equation}
\begin{equation} \label{eq2}
 \resizebox{.9\hsize}{!}{${f^*}(x_u) = f^*(x_l) + f^*(x_u) - f^*(x_l) = {f^*}(x_l) + \sum_{i=1}^{d} {w}^{*}_{i} (x_{u_i} - x_{l_i})$}
\end{equation}

We now attempt to answer the question of \emph{which} labeled point
to use. Here we make a classic machine learning assumption that the
model space used matches the data so that $f(x_l)$ matches $f^{*}(x_l)$.
Now we calculate the approximate error $\delta_{u}$ (due to not
having labels for the unlabeled data) for $x_u$ as the difference
between $f(x_u)$ in Eq. \ref{eq1} and ${f^*}(x_u)$ in Eq. \ref{eq2}:

\begin{equation} \label{eq3}
 \delta_u = |{f^*}(x_u) - f(x_u)| = |\sum_{i=1}^{d} ({w}^{*}_{i} - w_i) (x_{u_i} - x_{l_i})|
\end{equation}

We now propose an uncertainty measure $\theta$ for unlabeled point
$x_u$ and show that $\lambda \theta(x_u)$ upper bounds
$\delta_{u}$ in Theorem \ref{thm:bound}, where $\lambda$ is a constant.
\begin{equation} \label{eq4}
 \theta(x_u) = \min_{x_i \in L} L_1(x_u, x_i)
\end{equation}

Eq. \ref{eq4} means the current model's uncertainty for point $x_u$
is based on its nearest labeled point NN$(x_u)$ measured in $L_1$
distance. The following theorem shows the relationship between Eq.
\ref{eq3} (an unlabeled point's estimation error) and Eq. \ref{eq4}
(the $L_1$ distance between the unlabeled point and its nearest
labeled point).

\begin{theorem}
\label{thm:bound}
For any unlabeled point $x_u$, minimizing $\theta(x_u)$ is equivalent to minimizing the upper bound of estimation error $\delta_u$ for $x_u$.
\end{theorem}

Proof: From equation (\ref{eq3}) we have $ \delta_u = |\sum_{i=1}^{d} ({w}^{*}_{i} - w_i) (x_{u_i} - x_{l_i})|$.
\begin{equation} \label{eq5}
\begin{split}
\delta & = |\sum_{i=1}^{d} ({w}^{*}_{i} - w_i) (x_{u_i} - x_{l_i})|\\
& \le \sum_{i=1}^{d} |({w}^{*}_{i} - w_i)| \times |(x_{u_i} - x_{l_i})| \\
&\le \max_{i \in [1,d]} {|{w}^{*}_{i} - w_i|} \times \sum_{i=1}^{d}|(x_{u_i} - x_{l_i})|\\
& = \max_{i \in [1,d]} {|{w}^{*}_{i} - w_i|} \times L_1(x_u, x_l) = \lambda L_1(x_u, x_l)\\
\end{split}
\end{equation}
where $\lambda = \max_{i \in [1,d]}|w^{*}_i - w_i|$ is an unknown constant. 
Hence, the tightest upper bound for $\delta_u$ is achieved when $x_l$ is the nearest neighbor of $x_u$ in $L_1$ measure. 

 \begin{algorithm}[t]
 \small
 \caption{Algorithm for Sequential Active Learning}
 \label{alg:1}
 \begin{algorithmic}[1]
 \renewcommand{\algorithmicrequire}{\textbf{Input:}}
 \renewcommand{\algorithmicensure}{\textbf{Output:}}
 \REQUIRE \text{ }\\
 $L$: a set of labeled data ${{\{ (x_i, y_i) \}}^{l}_{i=1}}$.\\
 $U$: a set of unlabeled data ${{\{ x_j \}}^{n}_{j = l + 1}}$.\\
 $K$: the total query rounds.
 \REPEAT
 \FOR {each unlabeled $x_u$ in $U$}
 \FOR {each unlabeled $x_j$ in $U$}
 \STATE Calculate uncertainty $\theta(x_j)$ based on Eq. \ref{eq4}.
 \STATE Assume we have selected point $x_u$, calculate uncertainty $\theta^{x_u}(x_j)$ based on Eq. \ref{eq4}.
 \ENDFOR
 \STATE Calculate active selection value for $x_u$ via Eq. \ref{eq:Q_value}.
 \ENDFOR
 \STATE Select $x^*$ which maximize $Q$ as Eq. \ref{eq:selection}.
 \STATE Retrain the model with $(L \cup (x^*, y^*))$ and evaluate the model on the hold-out test set.
 \UNTIL Query round number reaches $K$.
  \vskip -0.15in
 \end{algorithmic}
 \end{algorithm}
 \setlength{\textfloatsep}{16pt}

\subsection{Proposed Active Selection Strategy for Sequential Query}
We now propose an evaluation function $Q$ for active selection which
chooses to query one point at each round. We first calculate the
current uncertainty for each unlabeled point $x_i \in U$ as
$\theta(x_i)$ based on Eq. \ref{eq4}. We then calculate the uncertainty
for each unlabeled point if $x_u$ had been queried, which we refer
to as ${\theta^{x_u}}(x_i)$. Next, we calculate the reduction of
uncertainty for each unlabeled point as $ \theta(x_i) -
{\theta^{x_u}}(x_i)$. Finally we calculate the difference between
the graphs defined in Section~\ref{prob:1} after querying $x_u$ as:
\begin{equation} \label{eq:Q_value}
 Q(x_u) = \sum_{i \in U} \theta(x_i) - \sum_{j \in U^{'}} \theta^{x_u}(x_j),
\end{equation}
where $U^{'} = U - \{x_u\}$. 
Now we select the point which maximizes the differences as $x^*$:
\begin{equation} \label{eq:selection}
 x^* = \argmax_{x_u} {Q}(x_u)
\end{equation}
The pseudocode for our proposed algorithm is summarized in Algorithm
\ref{alg:1}. At each iteration of active learning, our algorithm
selects a point $x^*$ which maximizes our active selection function.
After that, we retrain our model and test the newly trained model
with the hold-out test set. The whole process is repeated until the
number of querying rounds reaches the chosen maximum value. To speed up our proposed
algorithm, the outer loop's computations in Algorithm~\ref{alg:1} can be run
in parallel so the total time complexity for one query can be reduced
to $O(|U|*|L|)$.

 \begin{algorithm}[t]
 \small
 \caption{Algorithm for Batch Active Learning}
 \label{alg:2}
 \begin{algorithmic}[1]
 \renewcommand{\algorithmicrequire}{\textbf{Input:}}
 \renewcommand{\algorithmicensure}{\textbf{Output:}}
 \REQUIRE \text{ }\\
 $L$: a set of labeled data ${{\{ (x_i, y_i) \}}^{l}_{i=1}}$.\\
 $U$: a set of unlabeled data ${{\{ x_j \}}^{n}_{j = l + 1}}$.\\
 $K$: the total query points in a batch. \\
 $S$: the initial query set created by running Algorithm \ref{alg:1} for $K$ rounds.\\
 \REPEAT
 \FOR {each unlabeled $x_u$ in $U - S$}
 \FOR {each labeled $x_l$ in $S$}
  \STATE Based on Eq. \ref{eq:Q_value}, we now calculate the reduction for a set of points $S$ rather than a single point.
  \STATE Calculate the total uncertainty reduction $Q(S)$. 
  \STATE $S^{'} = S - \{x_l\} + \{x_u\}$
  \STATE Calculate the total uncertainty reduction $Q(S^{'})$. 
  \IF {$Q(S) < Q(S^{'})$}
  \STATE $ S = S^{'} $
 \ENDIF
 \ENDFOR
 \ENDFOR
 \UNTIL $\forall S^{'}$ $Q(S) > Q(S^{'})$.
 \STATE Output final query batch $S$.
 \end{algorithmic}
 \end{algorithm}
\setlength{\textfloatsep}{16.5pt}
\subsection{Proposed Active Selection Strategy for Batch Mode Query}

Most active learning methods have focused on sequential active learning
which selects a single point to query in each iteration. 
In this setting, the model has to be retrained after each new
example is queried and is not a realistic setting. (Multiple experts
annotate data in a parallel labeling system.) Moreover, using our
sequential active learning method to optimize the overall uncertainty
reduction in the bipartite graph is less accurate than optimizing
a group of points. We will demonstrate the performance advantage
of batch mode query in Section~\ref{sec:experimental}.

Our goal for batch mode query is to select a batch of $K$ data
points $S_u$ that provides the largest overall uncertainty reduction.
Searching for an optimal solution will be computationally expensive
(as suggested by the hardness results in Section \ref{sec:prob_def}).
To reduce the computation overhead, we propose to use a local search
approximation algorithm to optimize the batch mode query problem.

The local search algorithm \cite{arya2004local} with a single swap
is a popular algorithm for $k$-Median problem. We apply this idea
to our active learning strategy. Let the batch size be $K$. We
choose an initial set of $K$ query points using our sequential active
learning method (Algorithm \ref{alg:1}). We then repeatedly check the unlabeled
points in the pool and swap them with the $K$ query points if the
newly added point can reduce the total uncertainty reduction.
Reference \cite{arya2004local} proved that local search with $p$ swaps is a
$3 + 2/p$ approximation algorithm, which gives an approximation
ratio of $5$ in the single swap case with linear runtime. The
pseudocode for the batch mode query algorithm is summarized in
Algorithm \ref{alg:2}.

\subsection{Extension to Other Forms of Regression} 
Linear regression is popular for many real-world applications due
to its simplicity and interpretability. So far, we have discussed 
sequential and batch mode query strategies for linear regression.
However, it suffers from overfitting problem and 
does not generalize well to complex data sets. In this section, we
show how our current formulation can be extended to other forms of
regression.
To fit complex data, one common approach within machine learning
is to use linear models trained on non-linear functions of the
data. This approach maintains the generally fast performance and
high interpretability of linear models while allowing them to fit
a much wider range of data.

Let an instance $x$ with $d$ dimensions have
the coordinates $[x_1, x_2, ... x_d]$. We
construct the $k$-order polynomial features for $x$ by using the
multinomial theorem. Thus we have polynomial features like $x_i^k$
and interaction terms like $x_i * x_j^{k-1}$. A toy example for
second-order polynomial features with interaction terms for two
dimensional data is as follows: given the original data as: $x =
[x_1, x_2]$, the new feature is: $z = [x_1, x_2, x_1 x_2, x_1^2,
x_2^2]$ and the new regression model is: $f(z) = w_0 +
w_1 z_1 + w_2 z_2 + w_3 z_3 + w_4 z_4 + w_5 z_5$.

To prevent overfitting, we also impose the regularization term to
polynomial regression as ridge regression. By considering linear
fits within a higher-dimensional space built with these basis
functions, the model can be used for a much boarder range of
data. We also tested our approach for ridge regression. Due to limited space, our experimental results for ridge
regression appear in the supplementary material.

\section{Experiments}
\label{sec:experimental}
Here we aim to demonstrate the effectiveness of our approach empirically. 
In particular, we address the following three questions.
\begin{itemize}
 \item How do our results compare with baseline active learning methods for linear regression? This directly tests the tightness of our bound (Theorem \ref{thm:bound}).
 \item How does our proposed method perform when the query results are noisy?
 \item How does our proposed method perform on different forms of regression?
\end{itemize}
The first and second questions test the usefulness of our proposed
algorithms in practice while the last question addresses the
effectiveness of applying our proposed method to other forms of
regressions.

To compare the effectiveness of our proposed methods and answer the
above questions, our algorithms are compared with four representative
baselines.  
\begin{figure*}[tbh] \vskip -0.3in
 \subfloat[Housing]{
 \includegraphics[width=0.32\linewidth]{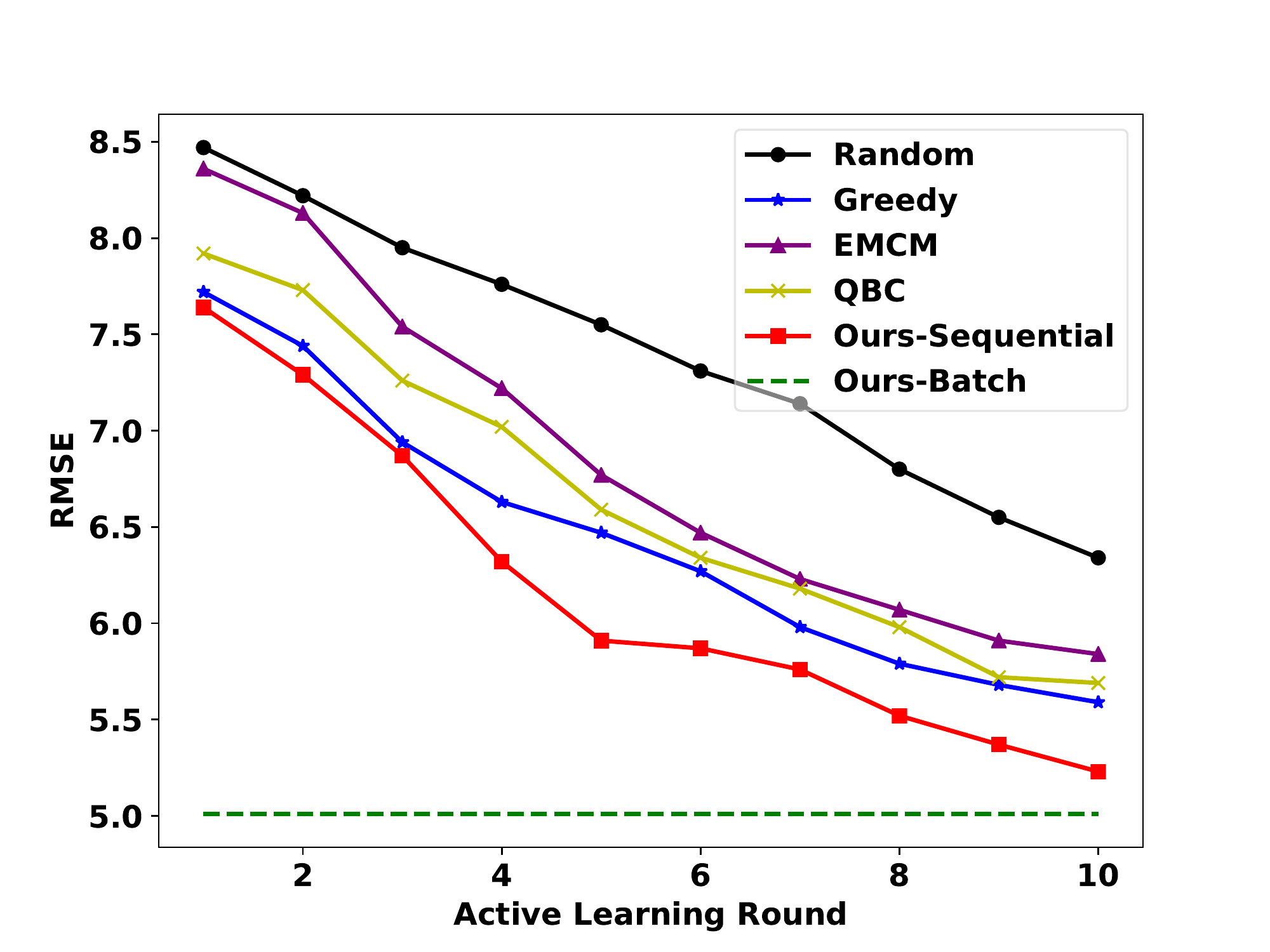}}
 \label{1a} \hfill
 \subfloat[Concrete]{
 \includegraphics[width=0.32\linewidth]{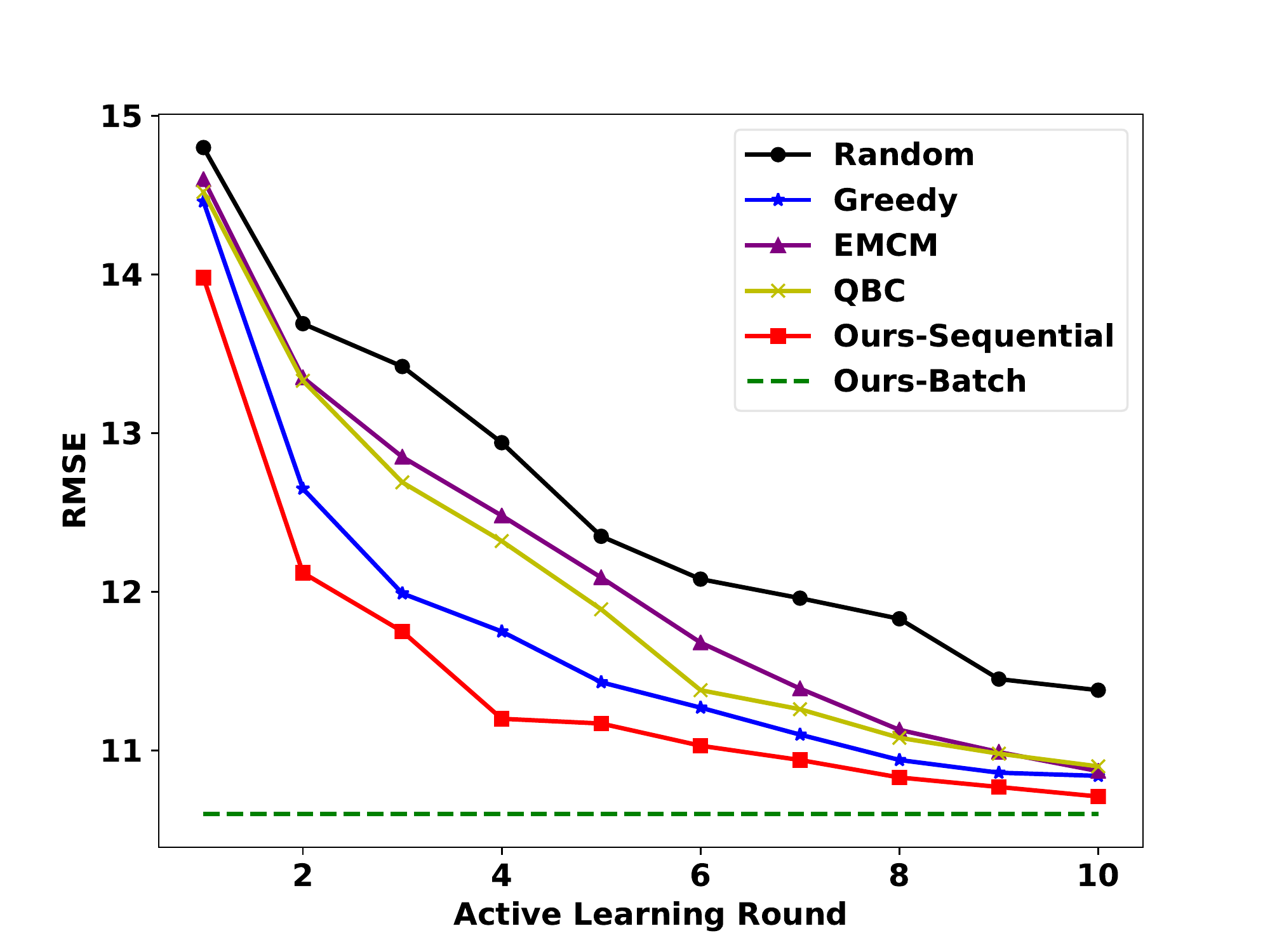}}
 \label{1b} \hfill
 \subfloat[Yacht]{
 \includegraphics[width=0.32\linewidth]{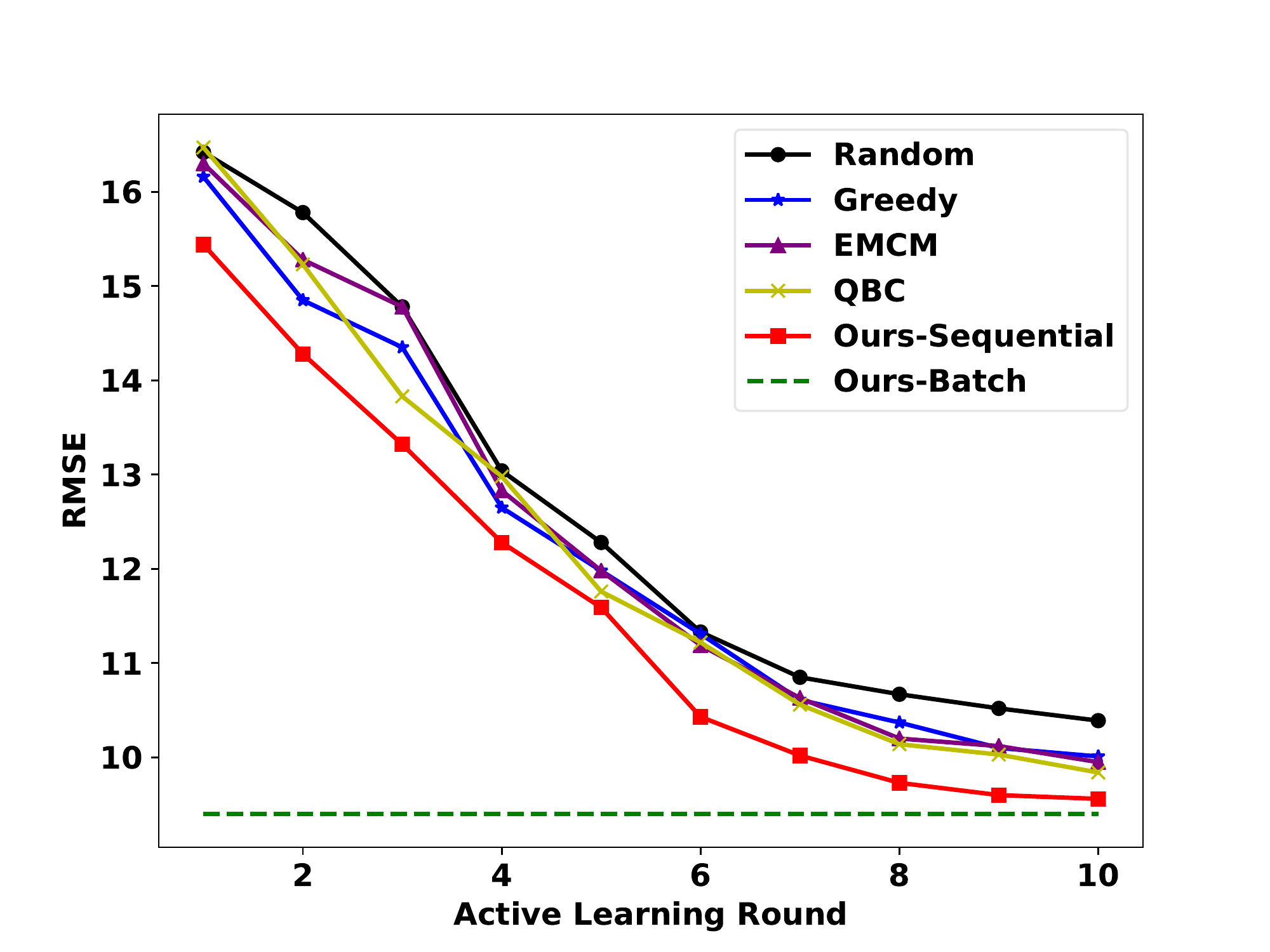}}
 \label{1c}
  \vskip -0.15in
 \subfloat[PM10]{ \includegraphics[width=0.32\linewidth]{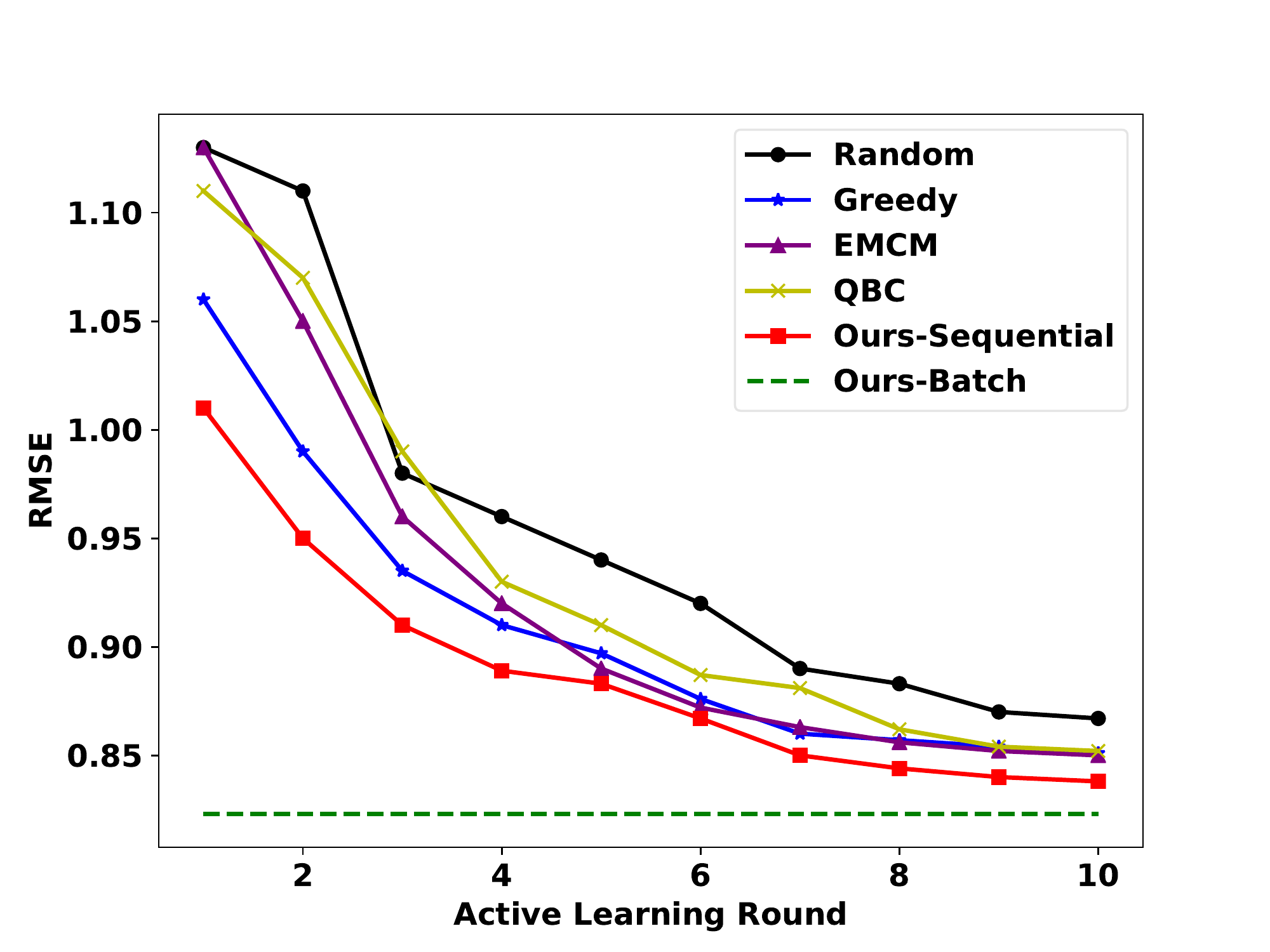}}
 \label{1d} \hfill
 \subfloat[Redwine]{
 \includegraphics[width=0.32\linewidth]{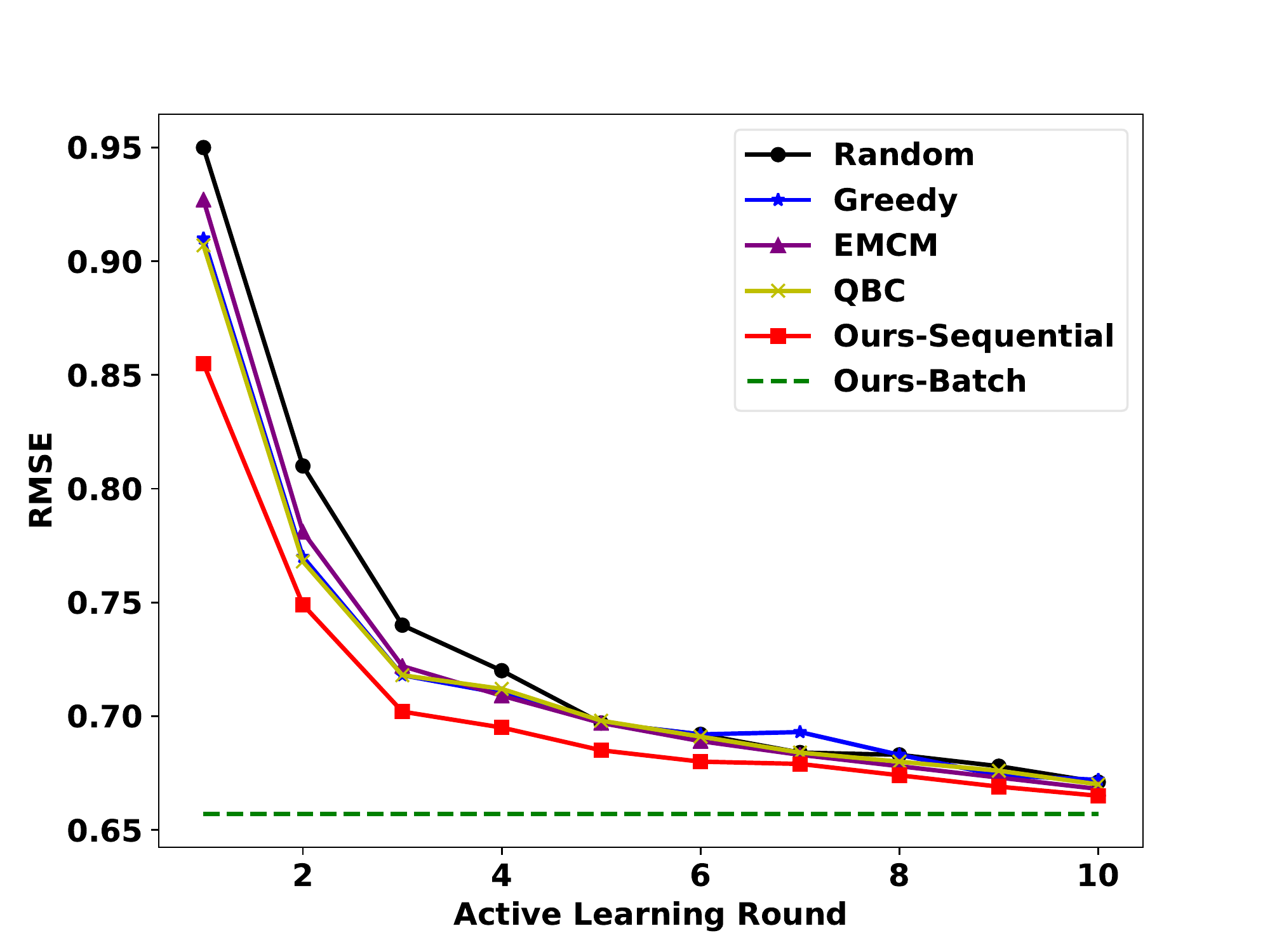}}
 \label{1e} \hfill
 \subfloat[Whitewine]{
 \includegraphics[width=0.32\linewidth]{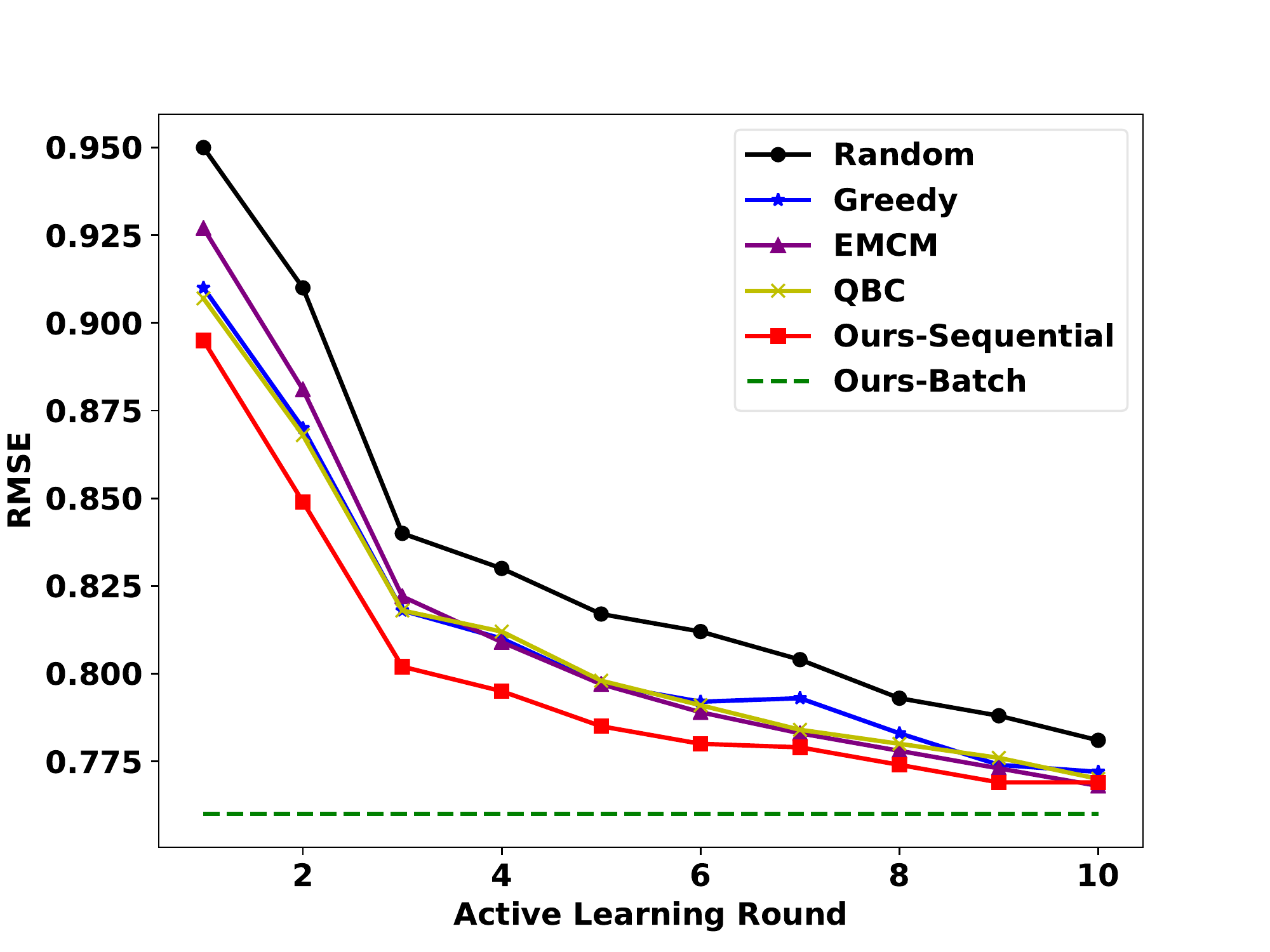}}
 \label{1f}
  \caption{Comparison results for the benchmark data sets with
  linear regression, (a) to (f) are learned with exact labels.
  Results are averaged over 30 random trials with ranking results
  shown in Table \ref{tab:ranking_table}.}
 \label{fig:performance_evaluation} \vskip -0.2in
\end{figure*} 
\begin{itemize}
 \item Random: Randomly select an instance from the unlabeled pool,
 which is widely used as active learning baseline.  
 \item Greedy passive sampling \cite{yu2010passive}: This approach will select
 an instance from the unlabeled pool which has the largest minimum
 Euclidean distance from the labeled set in feature space.  
\item
 Query by committee algorithm \cite{burbidge2007active}: This
 algorithm selects an instance that has the largest variance among
 the committee's prediction. The committee is constructed on bootstrap
 examples, and the number of committee members is set to be $4$.
 \item Expected model change maximization \cite{cai2013maximizing}:
 This algorithm quantifies the change as the difference between the
 current model parameters and the new model parameters, and chooses
 an unlabeled instance which results in the greatest change.
\end{itemize}

For all the baselines, we use the same parameters in original papers.
For simplicity, we use Random, Greedy, QBC, and EMCM to denote the
above baselines; our sequential query method is denoted as
Ours-Sequential, our batch query method is denoted as Ours-Batch.

\begin{table}[h]
\small
\vskip -0.05in
\caption{\small{Statistics and of Classical Experimental Data Sets Used in 
Active Regression (see \cite{yu2010passive} \cite{cai2013maximizing}).}}
\label{table:data_description}
\begin{center}
\vskip -0.2in
\begin{tabular}{c|c|c|c}
\toprule
\hline
Data sets & \# instances & \# features & Source\\
\hline
Housing & 506 & 13 & UCI\\
\hline
Concrete & 1030 & 8 &UCI\\
\hline
Yacht & 308& 6 &UCI\\
\hline
PM10 & 500 & 7&StatLib\\
\hline
Redwine & 1599 & 11&UCI\\
\hline
Whitewine & 4898 & 11&UCI\\
\hline
\bottomrule
\end{tabular}
\end{center}
\vskip -0.2in
\end{table}

\noindent
\textbf{Data Description.} We used six benchmark data sets which
are chosen from the UCI machine learning repository \cite{bache2013uci}
and CMU's StatLib \cite{vlachos2000statlib}: \emph{Housing, Concrete,
Yacht, PM10, Redwine, Whitewine}. These data sets were collected
from various domains and have been extensively used for testing
regression models. Their statistics and descriptions are shown in
Table~\ref{table:data_description}.

\noindent
\textbf{Experimental Configuration.} 
For each dataset, 1\% of the instances are sampled to initialize
the labeled set $L$. We use minimal initial data set to demonstrate
the advantages of our proposed algorithm. 
For evaluating the regression model for each run,
30\% of the instances are held out as the test set;
the rest of the instances are used for active learning.
For each run we will query $20\%$ of the unlabeled instances. Note
that we did not query all the unlabeled instances because the
performances of most methods converge after some queries. In
our plots we use query round as our x-axis; round one means $2\%$,
round two means $4\%$ of the unlabeled instances and so on; we have $10$
rounds in total. For the batch mode query, we will directly choose
a single batch consisting of all $20\%$ points to query.

We report the average results over the $30$ runs of experiments.
For the features in each dataset, we \emph{normalize} the features
using the standard score function (Z-score): $z = \frac{x-\mu}{\sigma}$
where $\mu$ is the mean of the population and $\sigma$ is the
standard deviation of the population. To simulate the noisy annotations
setting, we first calculate the standard deviation of the current
labels as $std(y)$ and then add a Gaussian noise $N(0, 0.1 * std(y))$
for each newly added query point.

\begin{figure*}[tbh]
\vskip -0.3in
 \subfloat[Housing$^*$]{
 \includegraphics[width=0.32\linewidth]{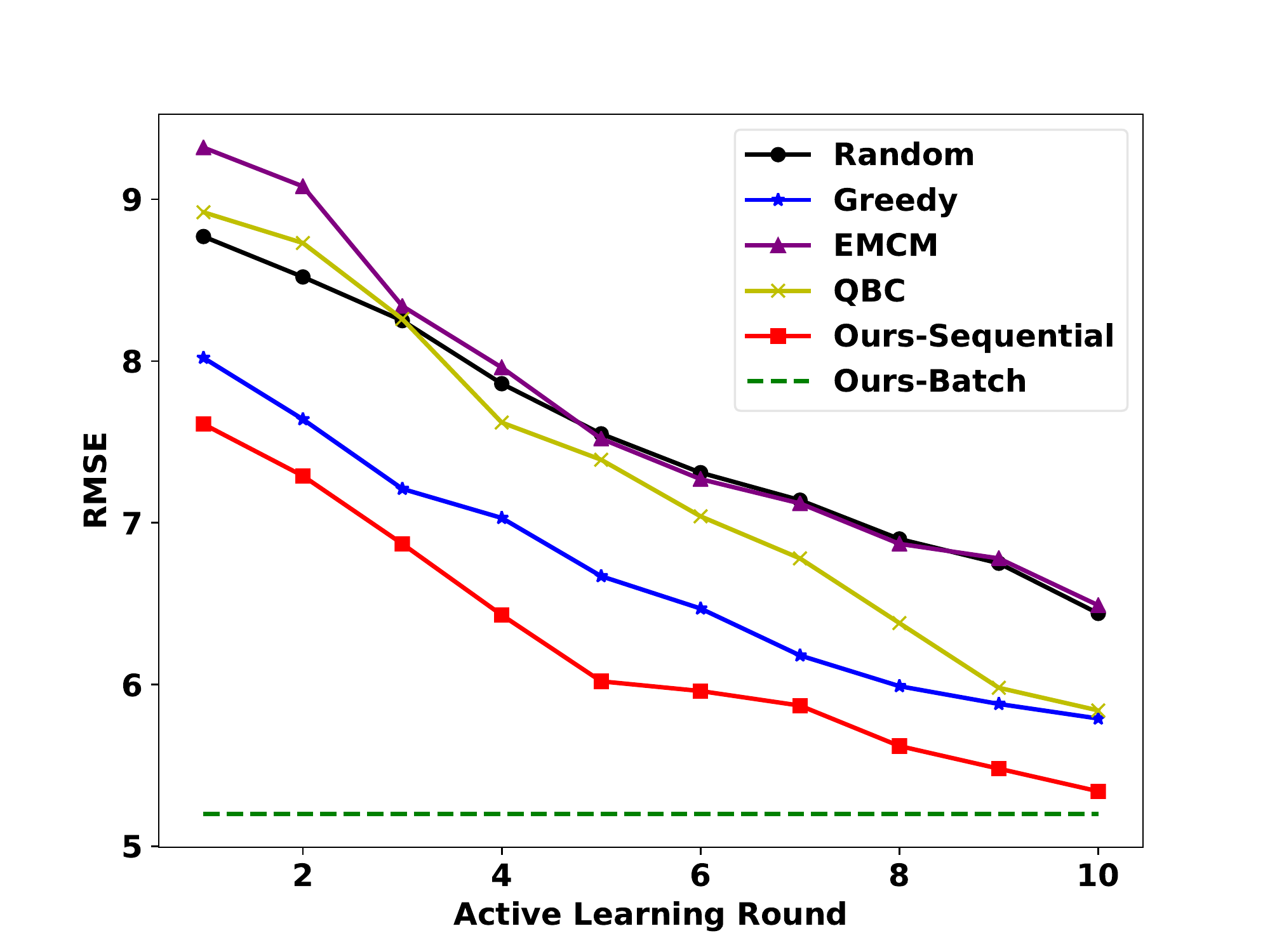}}
 \label{2a}
 \hfill
 \subfloat[Concrete$^*$]{
 \includegraphics[width=0.32\linewidth]{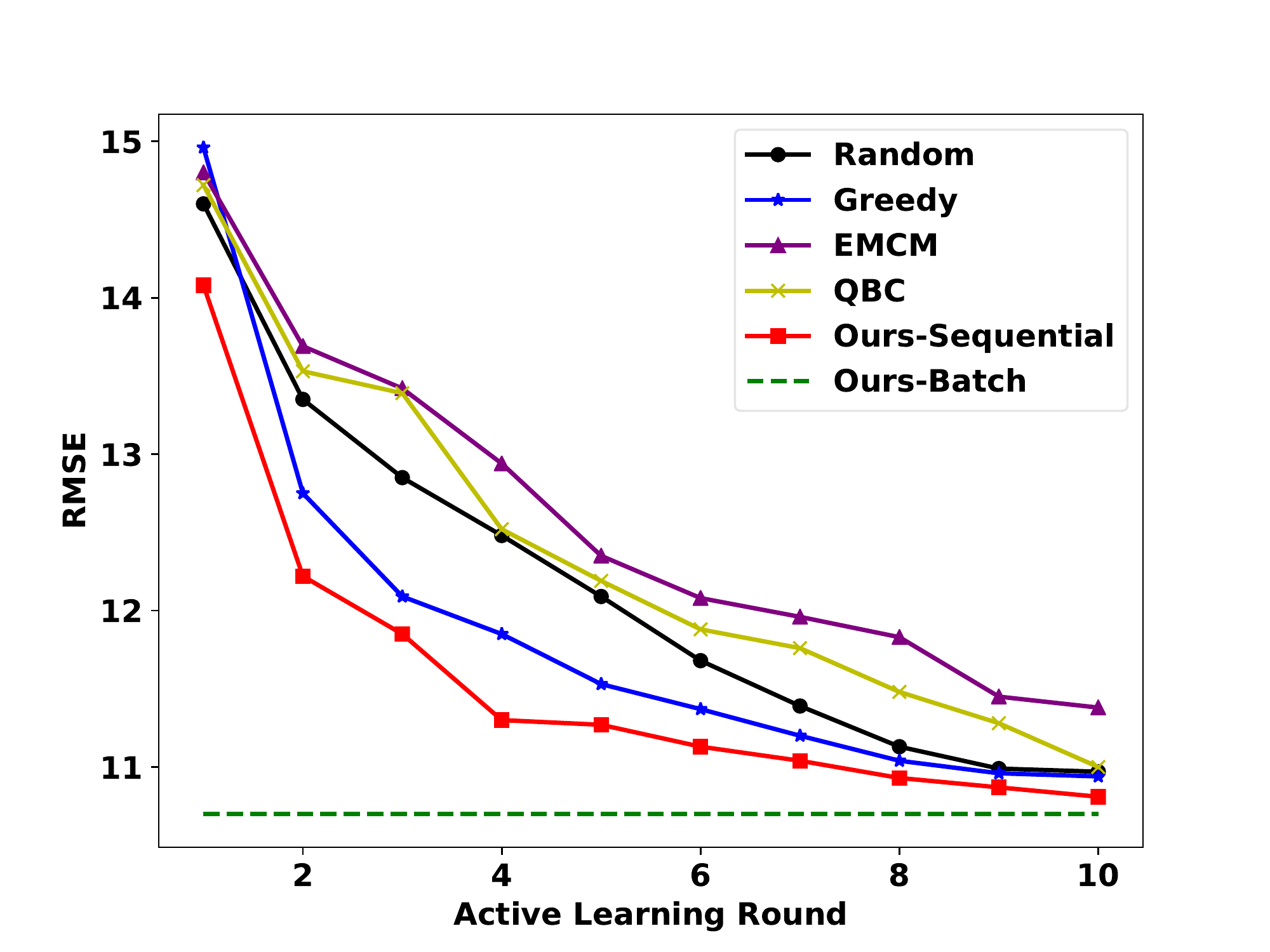}}
 \label{2b}
 \hfill
 \subfloat[Yacht$^*$]{
 \includegraphics[width=0.32\linewidth]{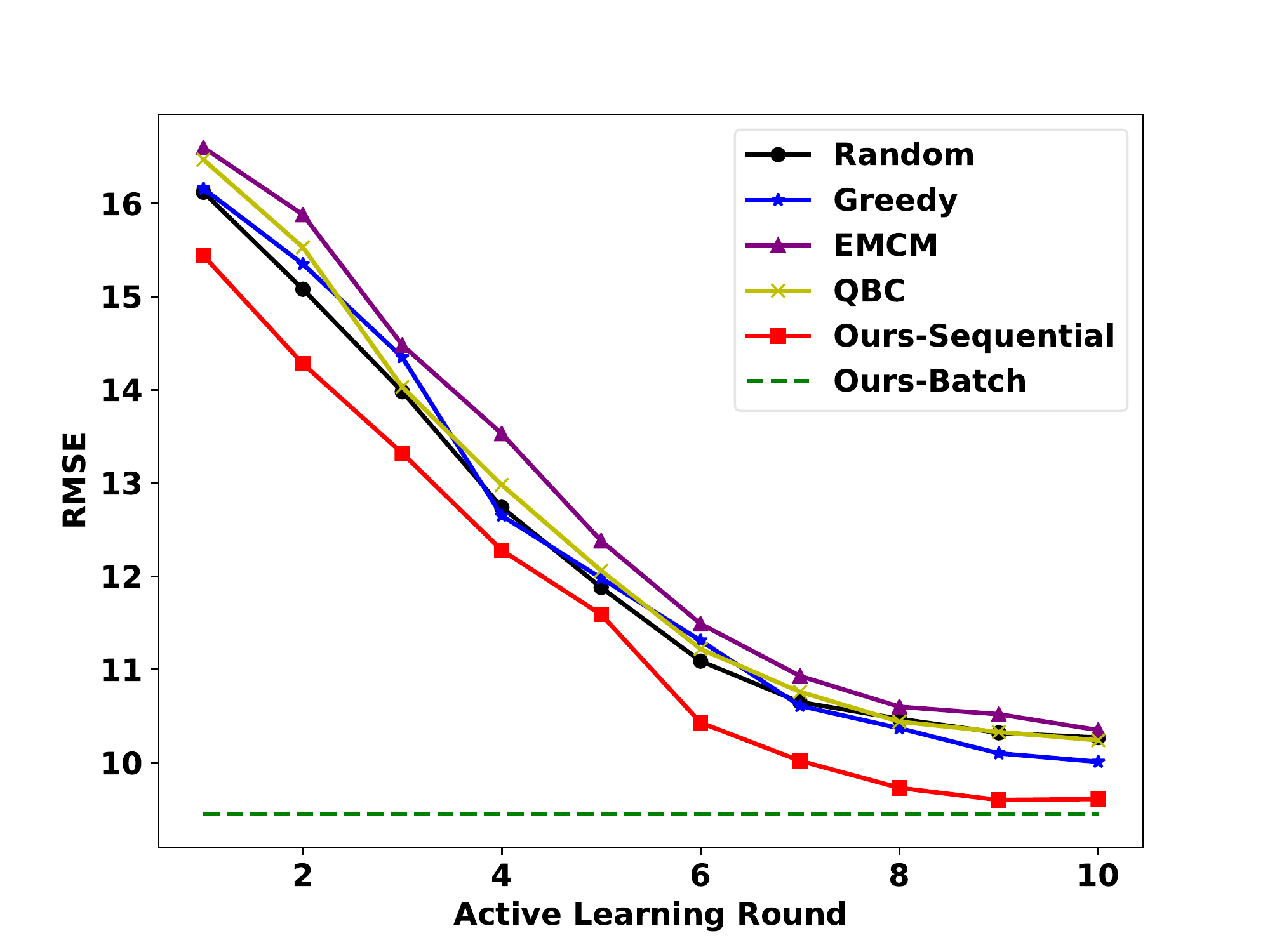}}
 \label{2c}
  \vskip -0.15in
  \subfloat[PM10$^*$]{
 \includegraphics[width=0.32\linewidth]{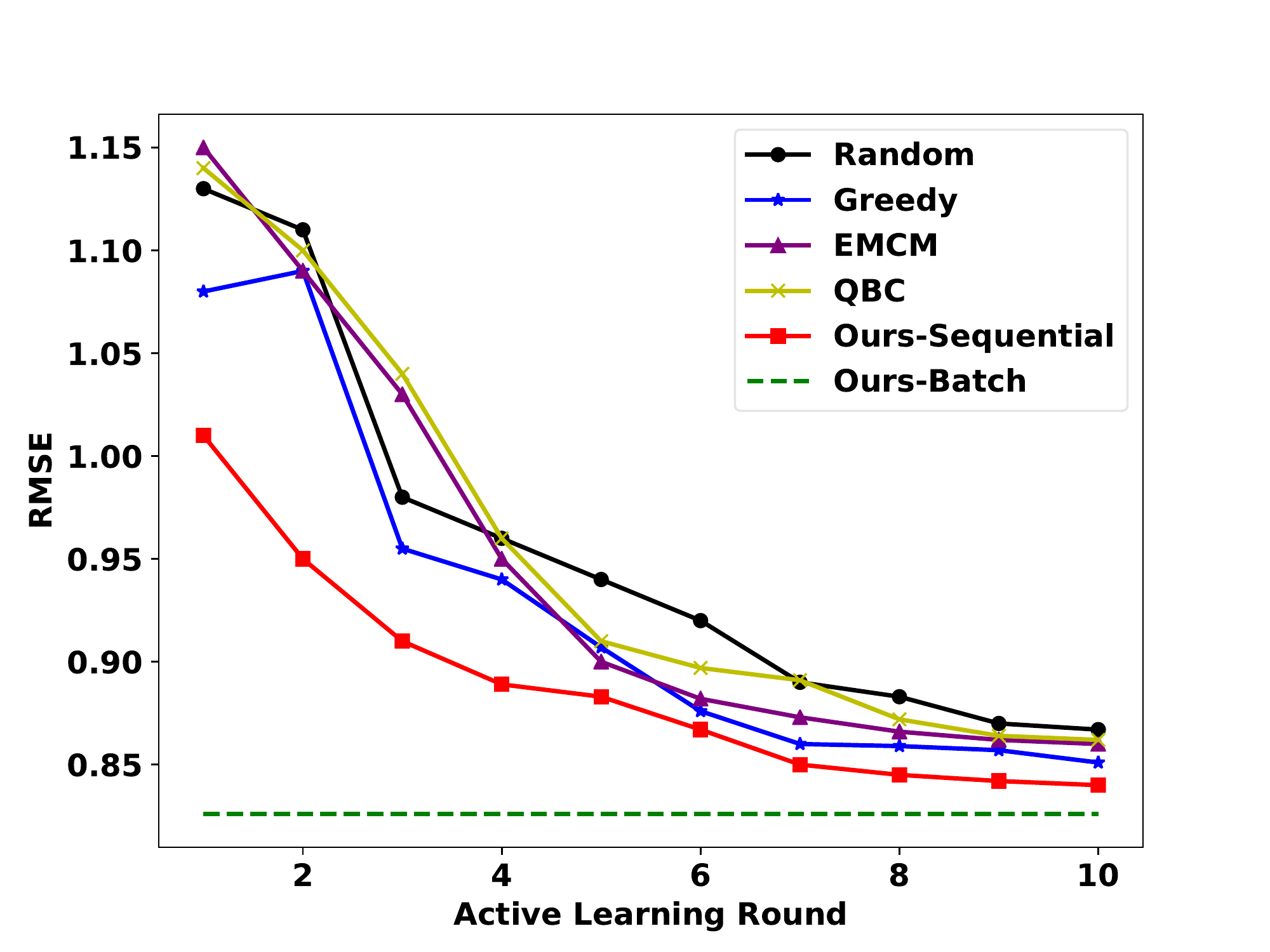}}
 \label{2d}
 \hfill
 \subfloat[Redwine$^*$]{
 \includegraphics[width=0.32\linewidth]{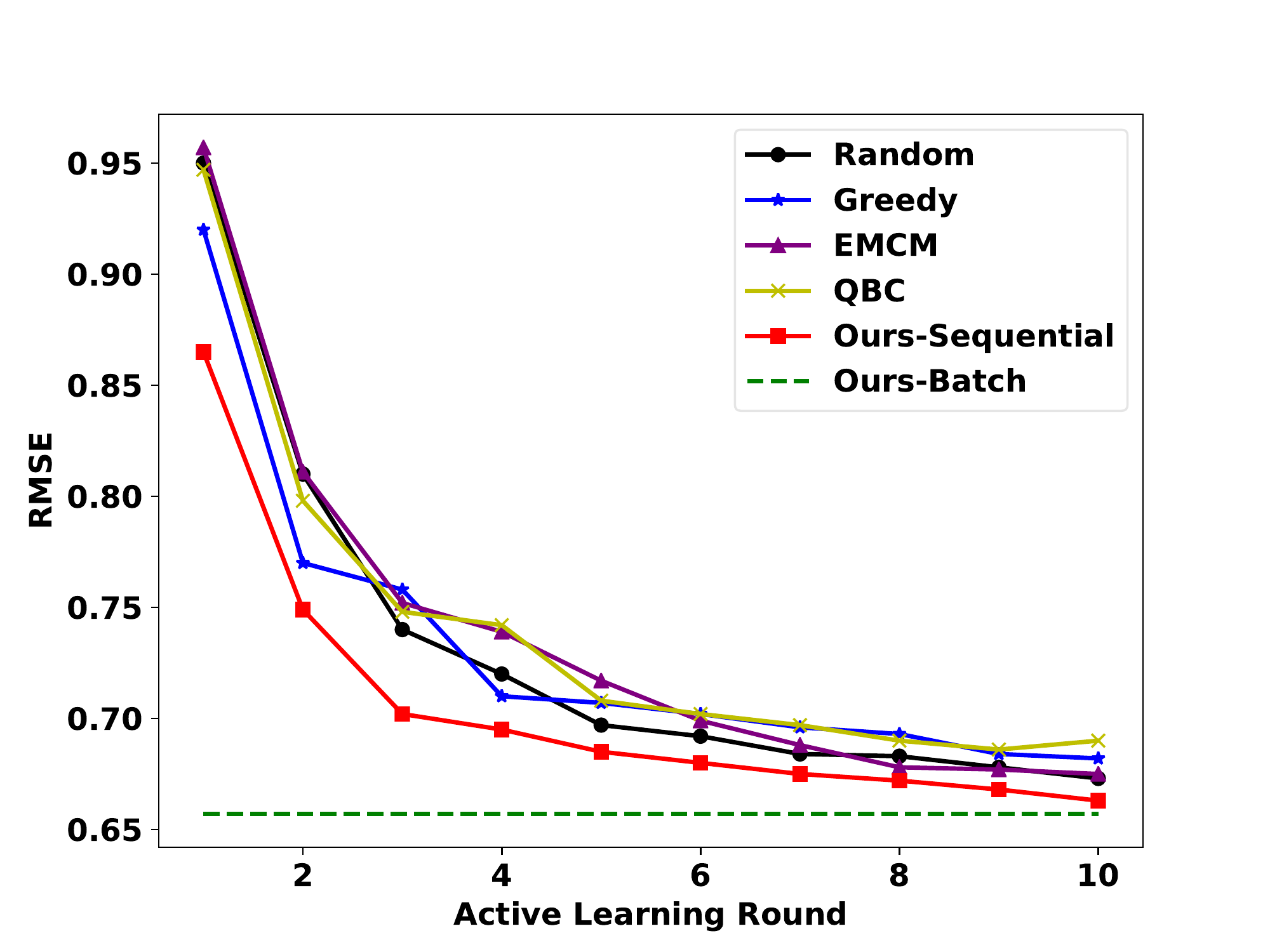}}
 \label{2e}
 \hfill
 \subfloat[Whitewine$^*$]{
 \includegraphics[width=0.32\linewidth]{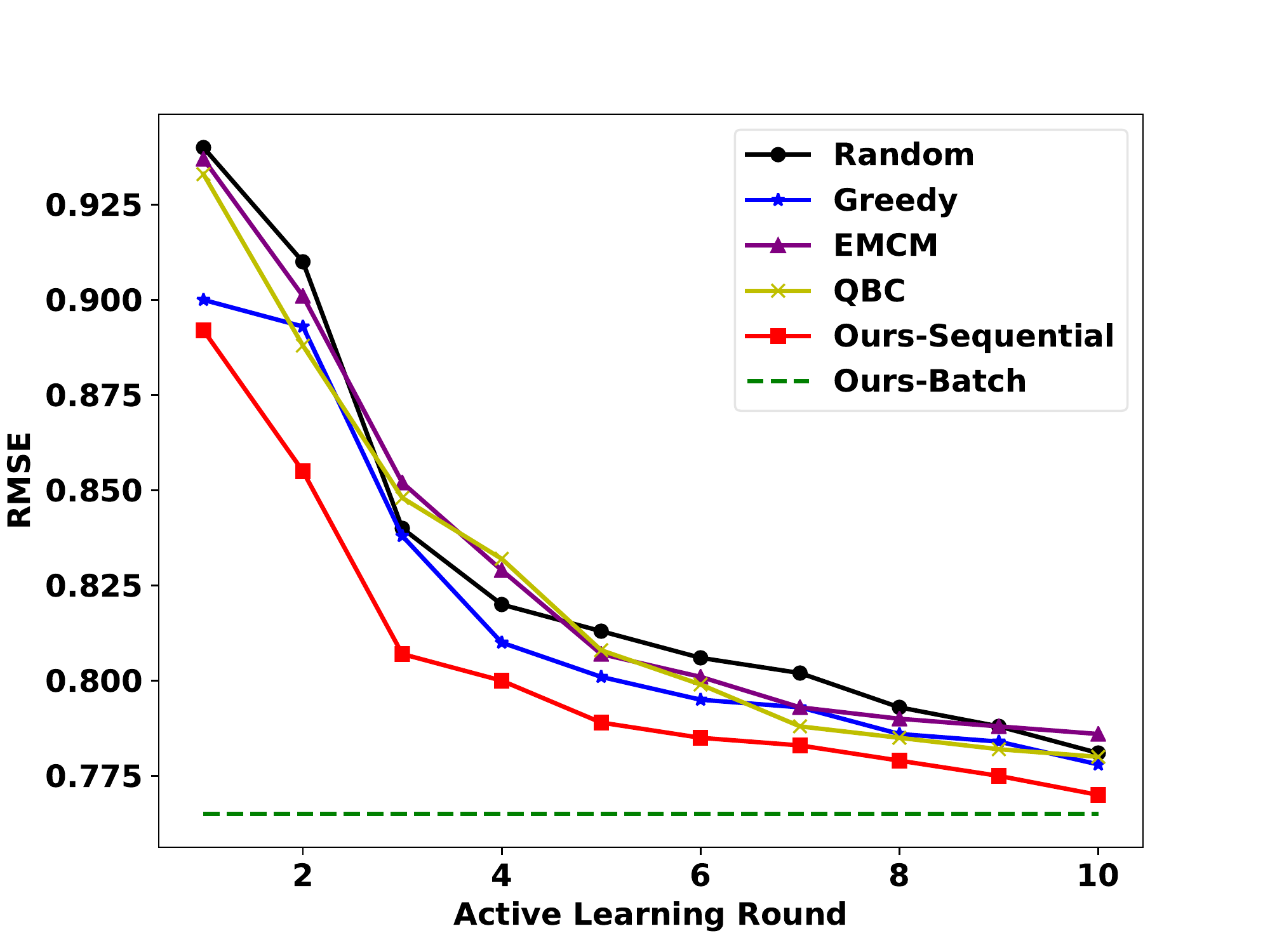}}
 \label{2f}

 \caption{Comparison results for the benchmark data sets with linear regression, (a) to (f) are learned with noisy labels. Results are averaged over 30 random trials.}
 \label{fig:performance_evaluation_noisy}
 \vskip -0.2in
\end{figure*}

\noindent
\textbf{Results for Non-Noisy Setting.}
Figure \ref{fig:performance_evaluation} plots the RMSE (Root Mean
Squared Error) curves against the total number of queries 
for all the compared approaches. We first discuss the noise-free
results in parts (a) to (f) of Figure~\ref{fig:performance_evaluation}.
Generally speaking, active learning methods
usually achieve lower RMSE than the Random method.
As can be seen from the plots, both QBC and EMCM perform better
than Random and their performances are close to each
other. This is reasonable because their ideas are similar; QBC
builds a committee to find the unlabeled point which has the largest
prediction variance while EMCM builds an ensemble to find the instance
which causes the largest changes in the model's parameters.

EMCM method tends to fluctuate at the beginning of the
training since its optimization objective only looks for unlabeled
instances that can bring the largest model change, regardless of the
the direction of change. Greedy sampling can achieve decent
performance in some of the data sets, but its performance is not
consistent; a possible reason is that different distributions
of the labeled and unlabeled data could fool the greedy method into
querying some outliers. \emph{Our proposed sequential query approach
consistently achieves lower RMSE than other methods in most cases.}

For the batch-mode query, one can see querying a batch is always
better than our sequential greedy query across all the data sets.
This result is expected since the local search approximation algorithm
reduces more in terms of the overall uncertainty for unlabeled
points. The better performance of batch mode query also reflects
that optimizing the overall uncertainty upper bound is useful in
practice. We haven't made comparisons to \emph{model-based batch
query algorithms} since their performance will worse compared
to their sequential query versions. For example, batch-EMCM performs
worse than sequential EMCM in \cite{cai2016batch}; this is because
the model is updated after each new example is chosen and added to the
training set so that each example is selected with more information.

\noindent
\textbf{Results for Noisy Setting.}
Parts (a) to (f) of Figure \ref{fig:performance_evaluation_noisy} show the RMSE curves
with noisy annotations. Both QBC and EMCM
perform worse than in the noise-free setting (see Figure
\ref{fig:performance_evaluation}). This result is expected because
we introduce Gaussian noise into each active query's feedback which
harms the performance of the \emph{model-based methods}. Both QBC
and EMCM assume the labels are noise-free and depend highly on the
quality of learned functions. The greedy method still performs
inconsistently, but it is less vulnerable to the noise because
it uses feature-based sampling strategy. Furthermore, our
proposed approach achieves lower average RMSE in a noisy setting.
\emph{This is to be expected as our method does not directly depend
on the regression function.}
\begin{table}[th!]
\small
\vskip -0.08in
\centering
\caption{\small{First place/second place/others counts of our method on
linear regression versus the baselines with varied numbers of
queries. The number of queries is represented by the percentage
of the unlabeled data. Each method is repeated for $30$ runs.}}
{\small
\tabcolsep=0.11cm
\begin{tabular}{ c|c|c|c|c }
\toprule
\hline
& \multicolumn{4}{|c}{Number of queries} \\
\cline{2-5}
&5\%& 10\% &15\%&20\% \\
\hline
Housing  &18/8/4 &18/9/3 &18/10/2 &18/10/2 \\
Concrete  &18/9/3 &19/9/2 &20/9/1 &20/9/1 \\
Yacht  &18/9/3 &18/10/2 &18/10/2 &18/10/2 \\ 
PM10  &15/12/3 &14/13/3 &14/13/3 &14/13/3 \\  
Redwine  &14/11/5 &14/11/5 &15/10/5 &15/10/5 \\ 
Whitewine &17/10/3 &17/10/3 &18/9/3 &18/9/3 \\
\hline
All  &100/59/21 &100/62/18 &103/61/16&103/61/16 \\
\hline
\bottomrule
\end{tabular}
\vskip -0.1in
}
\label{tab:ranking_table}
\end{table}

\begin{figure*}[tbh]
\vskip -0.3in
 \subfloat[Housing]{
 \includegraphics[width=0.32\linewidth]{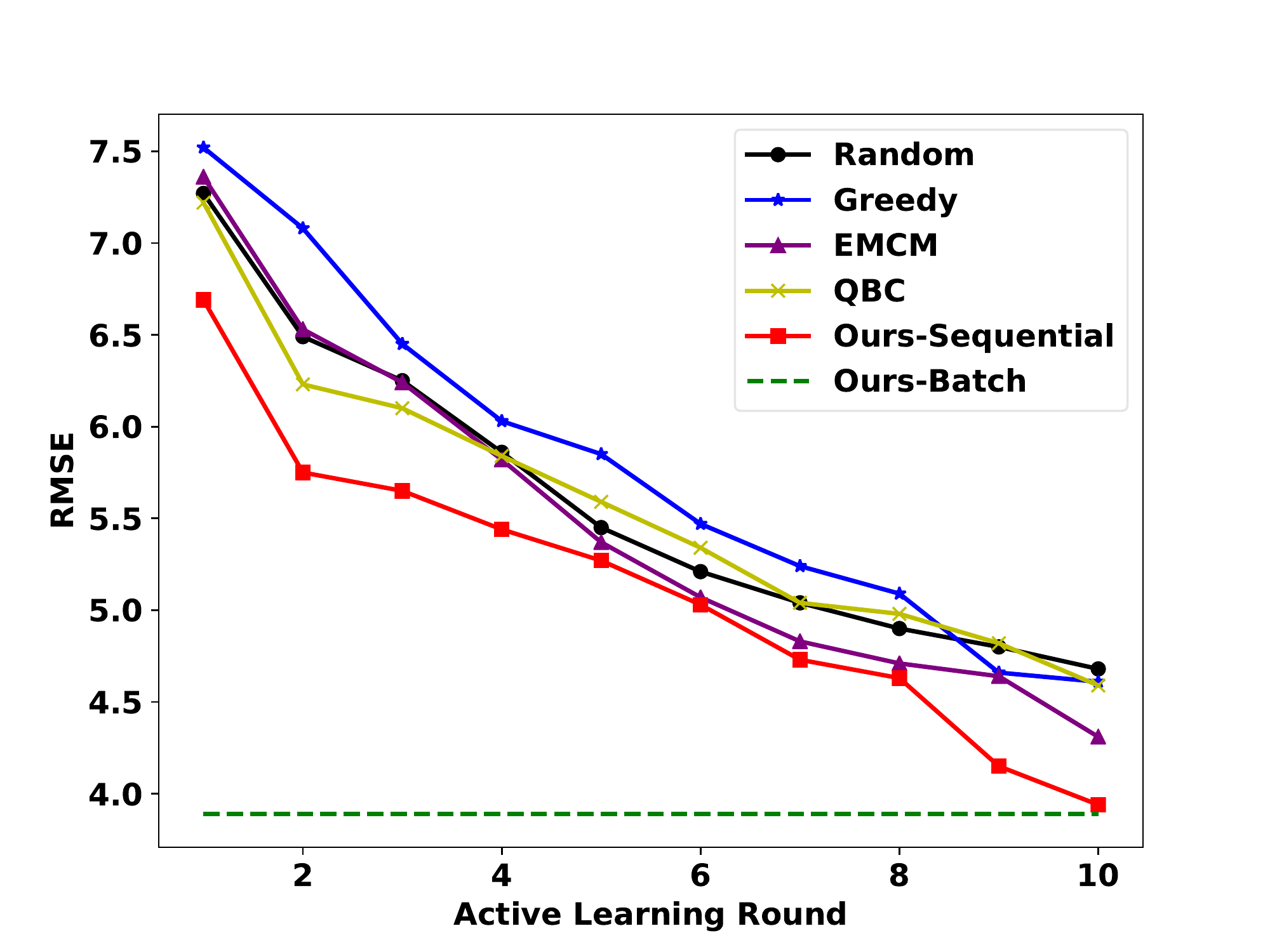}}
 \label{5a}
 \hfill
 \subfloat[Concrete]{
 \includegraphics[width=0.32\linewidth]{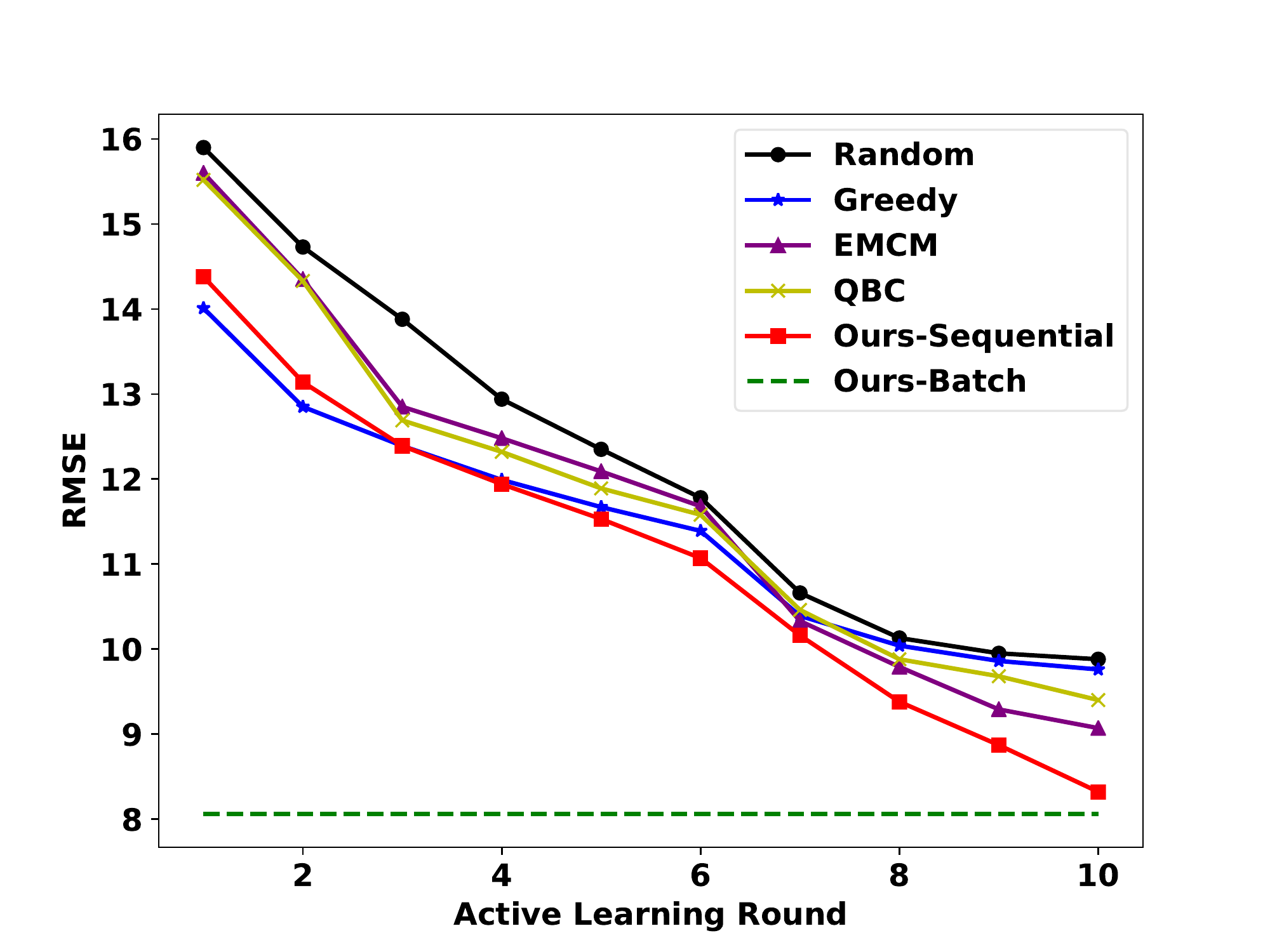}}
 \label{5b}
 \hfill
 \subfloat[Yacht]{
 \includegraphics[width=0.32\linewidth]{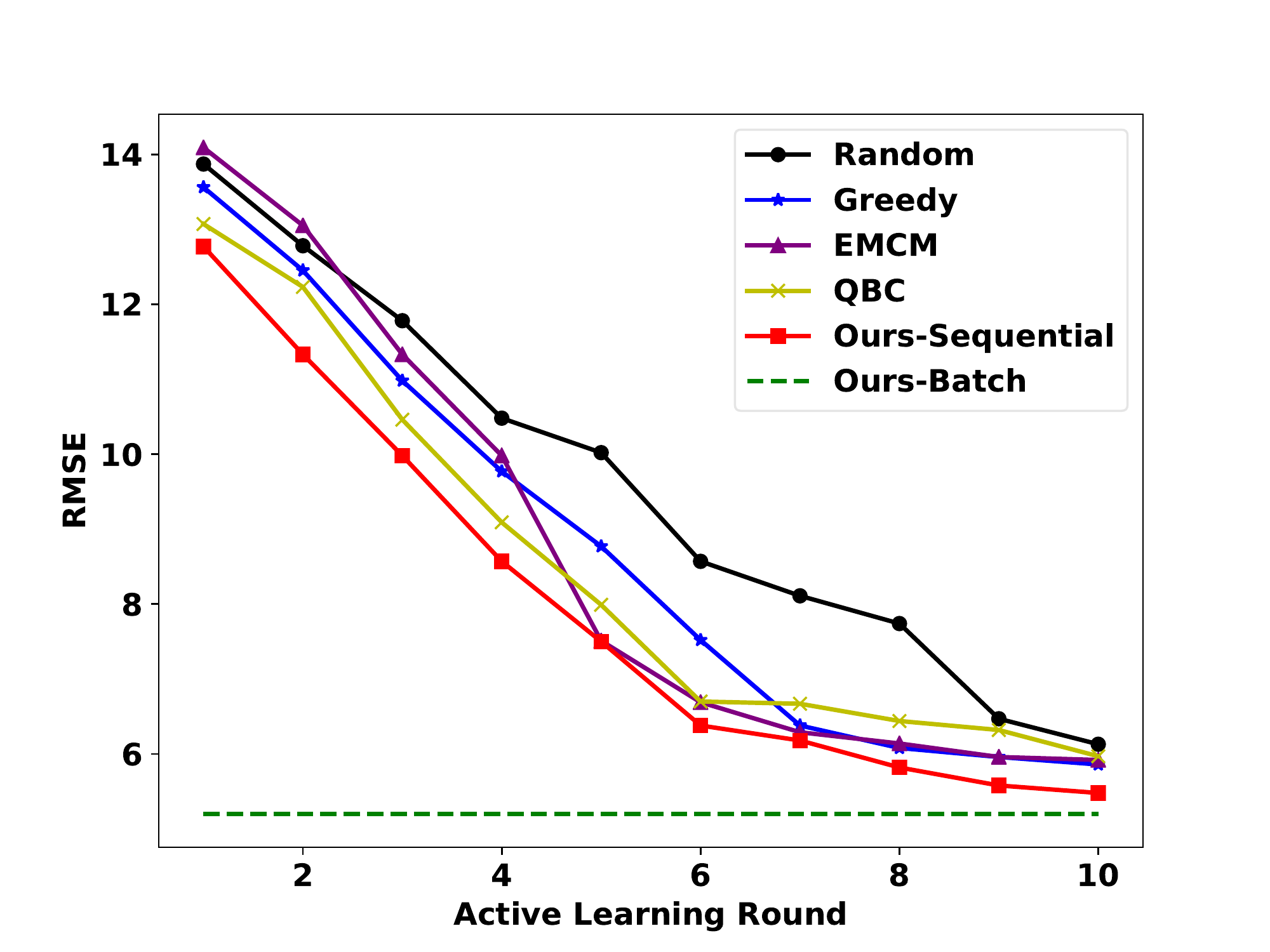}}
 \label{5c}
 \caption{Comparison results for the three harder (based on RMSE) data sets with polynomial regression. }
 \label{fig:polynomial_performance_evaluation}
  \vskip -0.1in
\end{figure*}

The overall performance of our sequential query method for
linear regression, shown in Table \ref{tab:ranking_table}, summarizes
the ranking counts of our method versus the other methods based on
the noise-free setting. Note that an active learning algorithm's
performance varies with the distributions of initial labeled and unlabeled
points. Thus, our method performs best in
more than half of the tests and behaves consistently as the number of
queries is increased. \emph{Significantly, this shows that our method does
not only performs better on average, but it also consistently outperforms
the competitors.}

\noindent
\textbf{Active Learning Experiments on Different Forms of Regression.}
We evaluated our extension for ridge regression and the results are
in line with the linear regression results. The details are summarized
in the supplementary material. To adapt to complex data, we discuss
experiments for polynomial regression in this section. We chose
three datasets, namely \emph{Housing}, \emph{Concrete,} and \emph{Yacht}. 
The reason for selecting them is that the performance of linear regression
for these datasets is not good enough. Figure \ref{fig:polynomial_performance_evaluation}
plots the RMSE curves for polynomial regression. We set the default
regularization parameter as $1$ across all the data sets.

Compared to linear regression results in Figure
\ref{fig:performance_evaluation}, we find that polynomial
regression works much better on these three datasets with a much
lower RMSE. Greedy behaves in a more unstable manner;
it achieves decent performance for \emph{Concrete} and \emph{Yacht} datasets but loses to Random for the \emph{Housing} dataset. This is possible because with larger feature dimensions, Greedy is more likely
to pick some outliers which are less informative. The EMCM and QBC
methods perform similarly across all data sets. They perform worse
at the beginning and gradually become better with an increase in the
number of labeled points.
Our proposed method performs consistently better than
the baseline methods, especially in the first $5$ active learning
rounds.

\section{Conclusions}
\label{sec:conclusion}
We propose a new graph-based approach for active learning in
regression. 
We formulate the active learning problem as a bipartite graph
optimization problem to maximize the overall reduction in uncertainty 
caused by moving points from the unlabeled collection to the labeled collection.
Experimental results on benchmark data show that the
proposed approach can efficiently find valuable points to improve
the active learning method. We explored both sequential and batch mode learning. Experimental results show that
the proposed measure of uncertainty and the method achieve promising
results for different forms of regression. 
A limitation of our method is
the high computational overhead
due to searching instances' neighbors in each active learning
round. 
In future work, we propose
to speed up our algorithms by using more efficient data structures
such as KD-trees and advanced hashing methods \cite{qian2013fast,
gilpin2013efficient}. Such techniques will enable us to apply our approach to large-scale regression tasks.

\section*{Acknowledgments}
We thank the SDM 2020 reviewers for providing helpful suggestions. 
This work was supported in part by NSF Grants IIS-1908530, OAC-1916805, IIS-1633028 and IIS-1910306.

\end{document}